\documentclass[final,3p]{elsarticle}
\usepackage{amsmath}
\usepackage[colorlinks=true,breaklinks=true,pdftex]{hyperref}
\usepackage{bbding}
\usepackage{pifont}
\biboptions{authoryear}
\usepackage{xcolor}

\usepackage{amsfonts}
\usepackage[utf8]{inputenc}
\usepackage{multirow} 
\DeclareUnicodeCharacter{202F}{\,}

\begin{document}

\begin{frontmatter}

\title{CMU-Flownet: Exploring Point Cloud Scene Flow Estimation \\ in Occluded Scenario}

\author{Jingze Chen$^{1}$, Junfeng Yao$^{*,1,2}$ , Qiqin Lin$^{1}$ , Lei Li$^{3}$}

\address{$^{1}$Center for Digital Media Computing,School of Film,School of Informatics, Xiamen University.\\
$^{2}$Institute of Artificial Intelligence, Xiamen University.\\
$^{3}$Department of Computer Science, University of Copenhagen
}


\begin{abstract}
Occlusions hinder point cloud frame alignment in LiDAR data, a challenge inadequately addressed by scene flow models tested mainly on occlusion-free datasets.
Attempts to integrate occlusion handling within networks often suffer accuracy issues due to two main limitations: a) the inadequate use of occlusion information, often merging it with flow estimation without an effective integration strategy, and b) reliance on distance-weighted upsampling that falls short in correcting occlusion-related errors. To address these challenges, we introduce the Correlation Matrix Upsampling Flownet (CMU-Flownet), incorporating an occlusion estimation module within its cost volume layer, alongside an Occlusion-aware Cost Volume (OCV) mechanism. Specifically, we propose an enhanced upsampling approach that expands the sensory field of the sampling process which integrates a Correlation Matrix designed to evaluate point-level similarity. Meanwhile, our model robustly integrates occlusion data within the context of scene flow, deploying this information strategically during the refinement phase of the flow estimation. The efficacy of this approach is demonstrated through subsequent experimental validation. Empirical assessments reveal that CMU-Flownet establishes state-of-the-art performance within the realms of occluded Flyingthings3D and KITTY datasets, surpassing previous methodologies across a majority of evaluated metrics.
\end{abstract}


\end{frontmatter}


\section{Introduction}
\begin{figure*}[!ht]

  \centering
  \centerline{\includegraphics[width=\textwidth]{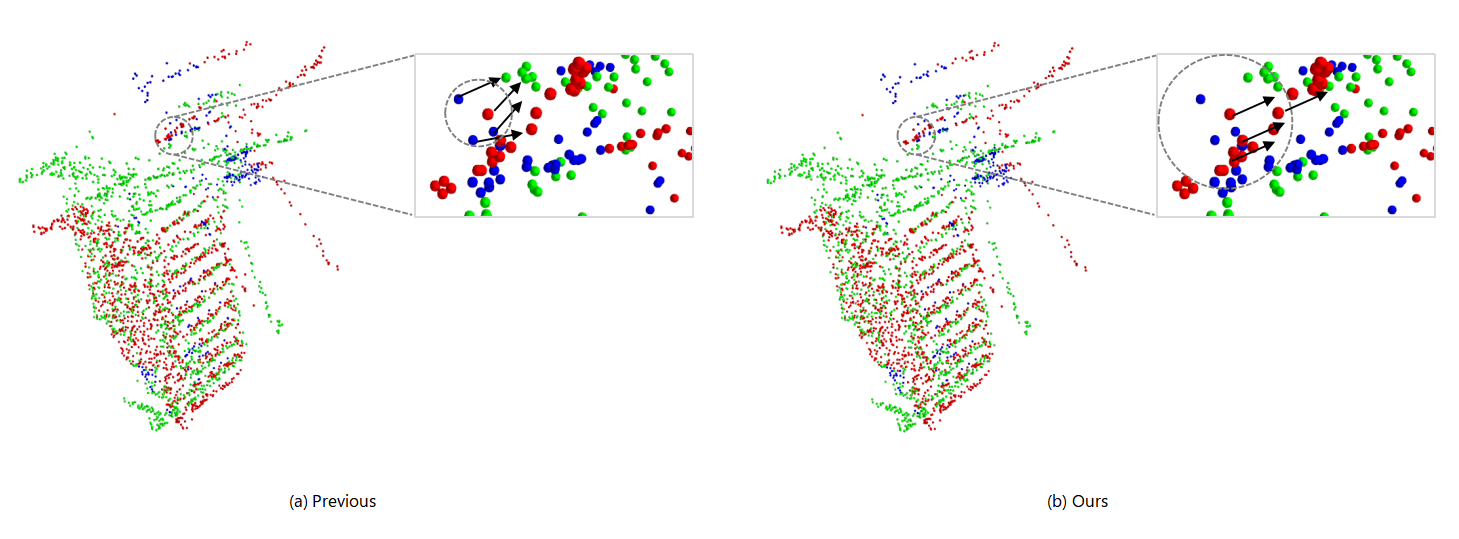}}
%

\caption{We illustrate the comparative methodology between conventional flow up-sampling techniques(left) and our proposed Matching Up-sampling framework(right), termed as CMU. Red points indicate positions at time $t$, green points represent positions at time $t+1$, and blue points signify areas that are occlued at time $t$. In contrast to previous methods that suffer from point mismatches due to their limited sampling scope, our CMU modules broaden the sampling range, and employs a correlation matrix to evaluate similarity at the point level to minimize errors.}
\label{fig:res}
\end{figure*}

The advent of deep neural networks promotes scene flow estimation methodologies. A big breakthrough was realized with the inception of FlowNet3D \cite{liu2019flownet3d}, a paradigm that harnessed the foundational principles of PointNet++ \cite{qi2017pointnet++} for the assimilation of local features into the fabric of scene flow estimation. This development marked the application of neural network architectures within this specialized domain. Subsequently, the development of HPLFlowNet \cite{Gu_Wang_Wu_Lee_Wang_2019} introduced an innovative mechanism for the computation of multi-scale correlations through the execution of upsampling operations embedded within bilateral convolutional layers. Building upon this, the work by \cite{Kittenplon_Eldar_Raviv_2020} unveiled a pioneering technique aimed at learning a singular iteration of an unrolled iterative alignment procedure, thus enhancing the precision of scene flow estimations. The introduction of 3DFlow \cite{wang2022matters} heralded a new epoch in the domain, establishing new benchmarks in terms of 3D End Point Error (EPE3D) and overall accuracy metrics.

Despite the substantial advancements achieved by these neural network architectures, the challenge of occlusion remains a significant impediment. In the context of point-to-point matching tasks, occlusions introduce a critical issue: points visible at a given moment $t$ may be obscured in the subsequent frame $t+1$ which may result in larger errors. In response to this pervasive challenge, Recent academic endeavors have concentrated on the integration of occlusion-aware mechanisms within the framework of neural network models, aiming to refine their understanding and processing of complex visual information by recognizing and accounting for occlusions. Zhao et al. \cite{zhao2020maskflownet} introduced an innovative asymmetric occlusion-aware feature matching module that is adept at learning a rudimentary occlusion mask. Such a mask is capable of filtering out regions rendered non-informative due to occlusion, immediately following feature warping processes. Further contributing to the discourse on occlusion mitigation, Saxena et al. \cite{saxena2019pwoc} unveiled a pioneering self-supervised strategy aimed at the prediction of occlusions directly from image data. This methodology represents a paradigm shift towards leveraging inherent image characteristics to infer occlusion patterns, thus facilitating a more nuanced and accurate scene interpretation.
 
The integration of occlusion estimation with scene flow estimation in point clouds represents a significant stride towards addressing the occlusion challenges in dynamic 3D scenes. Ouyang et al. \cite{ouyang2021occlusion} pioneered this approach by advocating for the exclusion of computed Cost Volume for points identified as occluded, thereby mitigating the detrimental effects of occlusions on scene flow accuracy. This methodology marked the inception of occlusion-aware scene flow estimation, illustrating the feasibility and importance of occlusion consideration within this domain. Building upon this foundation, Wang et al. \cite{wang2021festa} further advanced the field by proposing a network architecture that features novel spatial and temporal abstraction layers, both augmented with an attention mechanism. This architecture also integrates an occlusion prediction module, enhancing the network's ability to discern and appropriately account for occluded regions within the point cloud. This addition not only improves the robustness of scene flow estimation against occlusions but also paves the way for more sophisticated handling of temporal and spatial dynamics. In a further evolution of this domain, Zhai et al. \cite{zhai2023learning} introduced a cross-transformer model designed to capture more reliable dependencies between point pairs across frames. Integral to this model is the inclusion of occlusion predictions within both the network architecture and the loss function. 

The integration of occlusion estimation with flow prediction methodologies has indeed marked a forward leap in tackling the complexities of dynamic 3D scene analysis. However, a discernible performance dichotomy persists between occluded and non-occluded dataset evaluations. Upon a meticulous examination of contemporary algorithms, several critical limitations have been identified, contributing to performance inefficacy within occluded environments. Firstly, there exists an overarching deficiency in the comprehensive exploitation of occlusion information. This is primarily manifested in the prevalent approach of coupling occlusion estimation with flow prediction tasks, which often lacks a nuanced strategy for their integration. Predominantly, recent models, Wang et al. \cite{wang2021festa} and Zhai et al. \cite{zhai2023learning}, employ a multitasking framework that concurrently executes occlusion and flow prediction, yet may not fully leverage the potential synergies between these tasks. As noted in the studies by Cheng et al. \cite{cheng2022bi} and Wang et al. \cite{wang2022matters}, opt to overlook the information of occlusion, adopting a uniform treatment across all points within the scene. Moreover, while Ouyang et al. \cite{ouyang2021occlusion} innovatively apply occlusion data toward the refinement of Cost Volume feature extraction, the outcomes have yet to meet the anticipated benchmarks of efficacy. 

Otherwise, the acquisition of multi-scale point cloud features is a cornerstone in the development of advanced scene flow estimation models, with a prevalent reliance on a coarse-to-fine paradigm for both downsampling and upsampling processes, as demonstrated in seminal works by Wu et al. \cite{Wu_Wang_Li_Liu_Fuxin_2019}, Ouyang et al. \cite{ouyang2021occlusion}, and Zhao et al. \cite{zhao2020maskflownet}. This approach, while effective in a broad range of scenarios, predominantly utilizes a method of weighted upsampling based on Euclidean distances. Specifically, it involves the acquisition and weighted averaging of flow vectors from $K$ neighboring points, a technique predicated on spatial proximities. However, this prevailing strategy exhibits drawbacks in the context of occluded datasets, where the simplistic nature of the upsampling mechanism can inadvertently amalgamate the flow of occluded points with those of unoccluded points. This scenario underscores a critical limitation in the current methodology, whereby the simplistic Euclidean-based upsampling fails to discern between occluded and non-occluded points, leading to an elevation in error rates within occluded scenarios.

Based on the shortcomings of previous models, as Figure.\ref{fig:res} shown, we introduce the Correlation Matrix Upsampling Flownet (CMU-Flownet), a novel architecture which follows the coarse-to-fine paradigm. Our model incorporates an occlusion estimation module within its cost volume layer, alongside an Occlusion-aware Cost Volume (OCV) mechanism. Additionally, we propose an enhanced upsampling approach that expands the sensory field of the sampling process which integrate an Correlation Matrix designed to evaluate point-level similarity. Empirical evaluations demonstrate that CMU-Flownet sets a new benchmark for state-of-the-art performance in occluded Flyingthings3D and Kitti dataset. The key contributions of our study are outlined as follows:
\begin{itemize}
    \item We introduce a new Occlusion-aware Cost Volume (OCV) methodology to detect the occluded points and perform feature extraction, passing the cost volume containing the occlusion information to the flow prediction module.
  \item We propose Correlation Matrix Upsampling (CMU) module based on geometric structures and point features. This is a plug-and-play module that can be integrated into any flow prediction task. Experiments show that our up-sampling structure is more accurate than the traditional approach.
  \item Our method outperforms previous pyramidal structures on the occluded Flyingthings3d and Kitti datasets, further improving the performance of the neural network in occluded scenarios.
\end{itemize}

\section{Related Work}
\textbf{Scene Flow Estimation.} The concept of scene flow was first articulated by Vedula et al. in \cite{vedula1999three}. Scene flow, distinct from the 2D optical flow that delineates the movement trajectories of image pixels, is conceptualized as a vector characterizing the motion of three-dimensional objects. Early research in this field \cite{huguet2007variational, menze2015object,li2023hierarchical, vogel20113d,chen2023ssflownet} predominantly utilized RGB data. Notably, Huguet and Devernay \cite{huguet2007variational} introduced a variational approach to estimate scene flow from stereo sequences, while Vogel et al. \cite{vogel20113d} presented a piece-wise rigid scene model for 3D flow estimation. Menze and Geiger \cite{menze2015object} advanced the field by proposing an object-level scene flow estimation method, alongside introducing a dataset specifically for 3D scene flow. The advent of deep learning heralded transformative approaches in scene flow estimation. PointNet \cite{qi2017pointnet}, as a pioneering work, utilized convolutional operations for point cloud feature learning, which was further refined by PointNet++ \cite{qi2017pointnet++} through feature extraction from local domains. Subsequent studies \cite{liu2019flownet3d, Puy_Boulch_Marlet_2020, Wu_Wang_Li_Liu_Fuxin_2019,wang2022matters} have achieved impressive results in scene flow estimation. FlowNet3D \cite{liu2019flownet3d}, for instance, leverages PointNet++ \cite{qi2017pointnet++} for feature extraction and introduces a flow embedding layer to capture and propagate correlations between point clouds for flow estimation. Puy et al. \cite{Puy_Boulch_Marlet_2020} employed optimal transport for constructing point matches between sequences. Wu et al. \cite{Wu_Wang_Li_Liu_Fuxin_2019} proposed a cost volume module for processing large motions in 3D point clouds, while Wang et al. \cite{wang2022matters} innovated an all-to-all flow embedding layer with backward reliability validation to address consistency issues in initial scene flow estimation.

\textbf{Cost Volume.} 
The advent of Cost Volume as a transformative tool in the optical flow domain has catalyzed innovative developments in scene flow estimation. Wu et al. \cite{Wu_Wang_Li_Liu_Fuxin_2019} were at the forefront of this evolution, pioneering the integration of a Cost Volume module that predicts scene flow by constructing cost volumes at each level of the feature pyramid. To effectively accommodate large motions, the PointPWC-Net introduced a coarse-to-fine strategy, which involves concatenating features at level \(L\) with the upsampled features from level \(L+1\), thereby enhancing motion capture capabilities across different scales. Building upon this foundational pyramid structure, subsequent research endeavors have sought to refine and extend the utility of the cost volume concept. Wei et al. \cite{wei2021pv} proposed a groundbreaking approach that utilizes correlation volumes as a means to circumvent the limitations inherent in previous cost-volume based methodologies, specifically targeting the mitigation of error accumulation issues. Further, Cheng et al. \cite{cheng2022bi} drew inspiration from the upsampling and warping layers of PointPWC-Net \cite{Wu_Wang_Li_Liu_Fuxin_2019}, applying these techniques to enhance the fidelity of scene flow predictions.In a significant leap forward, Wang et al. \cite{wang2022matters} introduced an innovative all-to-all flow embedding layer, accompanied by a backward reliability validation mechanism. This approach is designed to tackle consistency challenges encountered in initial scene flow estimations, thereby setting new benchmarks for performance within the field at the time of its introduction.

\textbf{Occlusion in Flow Estimation.} 
The domain of optical flow estimation has witnessed significant advancements through the adept handling of occlusions. Drawing inspiration from these successes, Ouyang et al. \cite{ouyang2021occlusion} embarked on an innovative endeavor to harness occlusion data for enhancing Cost Volume feature extraction methodologies. Despite these efforts, the achieved outcomes have yet to fulfill the anticipated efficacy benchmarks. This has led to the incorporation of occlusion prediction modules within network architectures emerging as a pivotal strategy for augmenting accuracy in flow estimation tasks.Subsequent developments have seen scholars like Wang et al. \cite{wang2021festa} and Zhai et al. \cite{zhai2023learning} integrating this methodology into their models, thereby embedding occlusion prediction as a component of the loss function. Empirical evaluations of this approach have validated its effectiveness. This dual-faceted impact, wherein occlusion information for each point within the point cloud is ascertainable, coupled with the synergistic benefits of a multi-task fusion network architecture, fosters a conducive environment for the point convolutional layers. \cite{wang2022estimation} adopts a subnet to predict the occlusion mask and explicitly masks those occluded points, which ensures flow predictor to focus on estimating the motion flows of non-occluded points. \cite{lu2023gma3d} propose a module based on the transformer, which utilizes local and global semantic similarity to infer the motion information of occluded points.

\section{Problem Formulation}

Given two sequential point clouds of the identical scene, represented as $P = \{ (x_i, p_i) \in \mathbb{R}^{3} | i = 1,2, \ldots, n \}$ and $Q = \{ (y_j, q_j) \in \mathbb{R}^{3} | j = 1,2, \ldots, n \}$, where $x_i$ and $y_j$ are the coordinates of points in $P$ and $Q$ respectively, and $p_i$ and $q_j$ represent the feature attributes (such as color, normal vectors) at two different time frames. The objective is to compute a 3D motion field $F = \{ f_i \in \mathbb{R}^{3} | i = 1,2, \ldots, n \}$, which specifies the transformation vectors needed to align $P$ onto $Q$. This involves determining an optimal permutation matrix $M$ from the set $\{0,1\}^{n \times n}$ to satisfy the equation $P + F = MQ$, aiming to closely approximate the motion field $F$ to the ground truth $F_{gt}$ with high accuracy.

Concurrently, the analysis endeavors to ascertain the occlusion 
status $O(x_i)$ for each point $x_i$ originating from the first frame point cloud. Here, $O(x_i) = 1$ indicates that the point $x_i$ is unoccluded, while $O(x_i) = 0$ indicates occlusion. Distinguishing between occluded and non-occluded states is crucial for improving the accuracy of the 3D motion field prediction, which allows for the adaptation to dynamic occlusion scenarios that are common in sequential point cloud data.

\begin{figure}[h]
\centering
\includegraphics[width=0.9\linewidth]{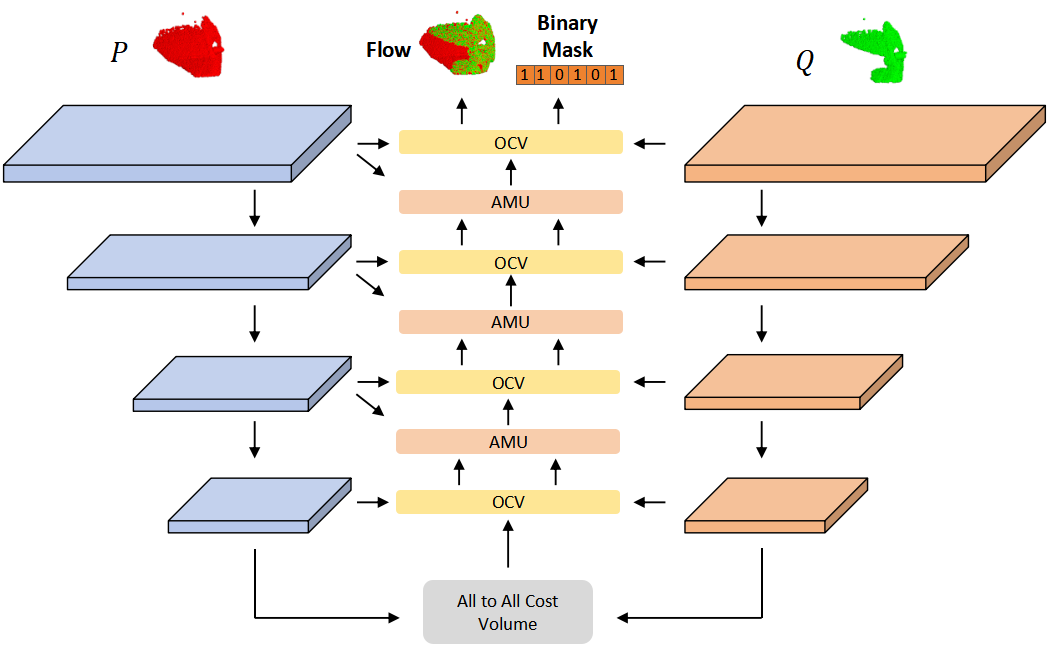}
\caption{We take the traditional pyramid-type structure as the overall framework of our model. Each frame of point cloud data is processed through a hierarchical point feature abstraction module consisting of four layers. Then a Correlation Matrix Upsampling (CMU) module is employed for upsampling purposes, and Occlusion-aware Cost Volume (OCV) is used for further refining the flow. After iterating through these processes for a specified number of loops, the model outputs the final predicted 3D motion field.}

\label{fig:network}
\end{figure}

\begin{figure*}[!ht]

  \centering
  \centerline{\includegraphics[width=\textwidth]{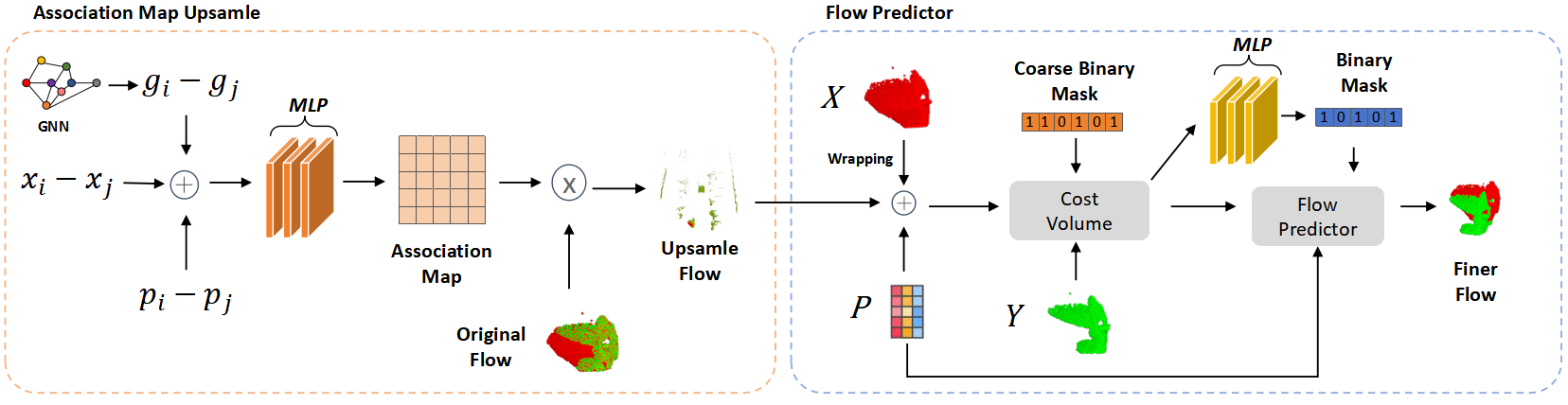}}
%

\caption{The picture shows the Correlation Matrix Upsampling (CMU) on the left and Flow Predictor on the right. Occlusion-aware Cost Volume (OCV) is the feature extractor in Flow Predictor. We use CMU for upsample flow and refine the flow by Flow Predictor.}
\label{fig:network_part}
\end{figure*}
\section{Method}
 We take the traditional pyramid-type structure as the overall framework of our model which is proven to be efficient in flow estimation task \cite{Wu_Wang_Li_Liu_Fuxin_2019,ouyang2021occlusion,wang2022matters}. We take $Q$ and $P$ as inputs.
Each frame of the point cloud data is processed through a hierarchical point feature abstraction module that consists of four layers. The abstraction process at each layer employs Farthest Point Sampling (FPS) for downsampling and uses PointConv \cite{qi2017pointnet++}. The spatial coordinates of the points at layer $l$ are denoted by $x^l$ and $y^l$. Feature inheritance is performed from the previous layer $l-1$, resulting in the derived features $p^l$ and $q^l$. This derivation involves operations such as grouping, pooling, and the application of weight-shared Multilayer Perceptrons (MLP). The sampling ratio at each layer is set to be 1/4 of the preceding layer, effectively reducing the number of points processed and refined at each subsequent stage. The overall architecture of CMU-Flownet is shown in \ref{fig:network}.

After extracting features from successive Pointconv layers, we adopt a coarse-to-fine paradigm to obtain scene flow at different scales progressively from one layer to the next. Motivated by \cite{wang2022matters}, we first apply the All-to-All Cost Volume at the bottom level of our network to build a correlation between two frames and learn the flow embedding. We then used the Correlation Matrix Upsampling (CMU) module for upsampling. We then add our proposed Occlusion-aware Cost Volume (OCV) to further refine the flow. Compared to the previous Cost Volume module, we add a masking prediction component and incorporate the masking information into the attention mechanism to obtain a more efficient cost volume. After a few loops output the final result, the predicted scene flow $F$ and the Binary mask $O(x_i)$.

In this section, we mainly discuss the proposed CMU and OCV modules. Detailed implementation information and schematic diagrams for all components discussed are provided in the supplementary materials.

\subsection{Correlation Matrix Upsampling}
Most point cloud processing methods follow a coarse-to-fine paradigm. Various upsampling strategies have been used to construct scene flow fields from sparse levels to dense levels with proper weights. some work like \cite{liu2019flownet3d,Wu_Wang_Li_Liu_Fuxin_2019,ouyang2021occlusion} et. al. predict flow based on trilinear interpolation upsampling, it is simple and efficient, but can lead to error accumulation, especially in occluded datasets. \cite{gu2019hplflownet,wang2021hierarchical} proposed an intra-frame patch features based method which uses interpolation functions to represent the distances between each point and its neighboring points. The proposal of this method has improved some accuracy, but it is still unable to adapt to complex occlusion Scenario and larger sensory field. Inspired by \cite{shen2023self} which use features to update superpoints, we productively propose an enhanced upsampling approach that expands the sensory field of the sampling process which integrate an Correlation Matrix designed to evaluate point-level similarity.

Our method attempts to generate an upsampling flow that satisfies the following requirements: (1) Neighbouring points are with similar flow patterns; (2) Points with similar characteristics share similar flow. Thus, we introduce graph-structure based coding on the point cloud to extract point cloud features. First, we construct the graph structure at the point level, where the edges of the graph aggregate information such as the position of the points, colour and normal vector. Then we use setconv layer as encoder to learn neighbourhood information, Combine the previously obtained features $p^l$ and $q^l$. Subsequently, we use MLP to learn the similarity of different point features.

\subsubsection{Flow-Graph Encoder}
Graph Neural Networks (GNNs) are a category of neural networks designed specifically for processing data structured as graphs. Graphs are mathematical structures used to model pairwise relations between objects, characterized by vertices (nodes) and edges (links). Work in \cite{shi2020point} demonstrates the effectiveness of GNN in point cloud processing tasks.

We take the upsampling process from $l+1$ to $l$ layer as an example to illustrate the Correlation Matrix calculation. Next, we define the graph structure as follows:
\begin{equation}
   \begin{aligned}
       E^l = \{(x^{l}_i-x^{l+1}_j||p^{l}_i||p^{l+1}_j)~|~\|x^{l}_i-x^{l+1}_j\|_2<r\}
   \end{aligned}
   \label{eq1}
\end{equation}
Equation \ref{eq1} represents the establishment of the graph structure. In Equation \ref{eq1}, $E$ denotes the edge set in the graph structure. We select the neighboring points around each point and use their distance and feature differences as the criteria for edge formation. \(p^{l+1}_i\) and \(p^l_j\) denote the feature in two different layers.  $||$ is the contact operator. Following the methodology suggested by Kittenplon et al. \cite{Kittenplon_Eldar_Raviv_2020}, we encode edge features using three consecutive $setconv$ layers as our convolution mechanism.

In contrast to previous GNN methods, we introduce the concept of spatial memory in the feature extractor. Spatial memory has been effectively applied in the field of semantic segmentation, where studies such as \cite{tokmakov2017learning,nilsson2018semantic} demonstrate that sequential input outperforms single-frame input by enabling the neural network to incorporate temporal information. While methods by \cite{Wu_Wang_Li_Liu_Fuxin_2019, ouyang2021occlusion} utilize wrapped points to gather neighborhood information across different frames, our model diverges by leveraging this temporal and spatial information specifically for flow upsampling. 
In the context of scene flow estimation, which inherently carries temporal data, our model attaches the coarse flow to point $P$ at time $t-1$ and retains memory up to point $Q$ at time $t$, thus facilitating the learning of geometric information across time intervals. 

\begin{equation}
    \begin{aligned}
        &x^{l}_{w,i} = x^{l}_{i}+f^{l}_{c,i} \\
        E^l_w = \{(x^{l}_{w,i}&-y^{l+1}_j||p^{l}_i||q^{l+1}_j)~|~\|x^{l}_{w,i} - y^{l+1}_j\|_2<r\}
        \label{eq2}
    \end{aligned}
\end{equation}

At the beginning, we take the traditional approach to get the initial coarse flow \(f^{l-1}_{c,i}\). And $x_{w,i}$ denotes the $i^{th}$ point in $P$ wrapped by coarse flow, $E^l_w$ indicates the features of point $p_{w,i}$ with the memory module, and we keep points to the next frame to learn the features of the surrounding points. We use $setconv$ layers to encode the features. $g$ and $g_{w,i}$ denotes the finished encoded feature. As previously articulated in the formula, $g^l_{w,i}$ denotes the feature which is obtained from wrapped points.
\begin{equation}
    \begin{aligned}
        &g^l_i = setconv(e^l_i),~e_i \in E^l \\
        &g^l_{w,i} = setconv(e^l_{w,i}),~e^l_{w,i} \in E^l_w
        \label{eq3}
    \end{aligned}
\end{equation}

\subsubsection{Correlation Matrix}
Flow consistency algorithms based on distance only can lead to prediction errors because the motion patterns of object boundary points are very different from those of surrounding points. So we model based on Euclidean distances and feature distances, the purpose of which is to reduce errors. Let $I$ denotes the total number of points in layer $l$, and $N$ denotes the $N$ nearest neighbouring points around $x^l_i$ (We set $N = 32$). Correlation Matrix is a module for learning point cloud similarity based on a feature encoder, where we combine the temporal and spatial features learned in the previous section to generate an $I \times N$ matrix that measures the similarity between point levels.

\begin{equation}
    \begin{aligned}
    &u_{i,n} = (x^{l}_i||x^{l}_{w,i}) - (x^{l+1}_n||x^{l+1}_{w,n})\\
    &v_{i,n} = (g^{l}_i||g^{l}_{w,i}) - (g^{l+1}_n||g^{l+1}_{w,n})\\
    &w_{i,n} = (p^l_i||p^l_{w,i}) - (p^{l+1}_n||p^{l+1}_{w,n})\\
    a_{i,n} &= MLP(u_{i,n})+MLP(v_{i,n}) + MLP(w_{i,n})
    \end{aligned}
    \label{eq4}
\end{equation}
Where $a_{i,n}$ denotes the degree of similarity between the $i^{th}$ layer $l$ point and the $n^{th}$ layer $l+1$ point. Next, we assign each point $p_i$ a similarity vector, We map the similarity parameter to the interval $[0, 1]$ as a weight for flow upsampling.
\begin{equation}
\begin{aligned}
   a_{i,n} &= softmax([a_{i,1}, a_{i,2}..., a_{i,N}])_n
    \label{eq5} 
\end{aligned}
\end{equation}

We update the scene flow vector with the Correlation Matrix that maps the ground truth labels to each point, $f^l_{u,i}$ means the $i^{th}$ upsampling flow.
\begin{equation}
    f^{l}_{u,i} = \sum^{N}_{n=1} a_{i,n}* f^{l+1}_n
    \label{eq6}
\end{equation}
The inclusion of correlation and weighted summation ensures that the upsampling process is both context-aware and spatially precise, thereby improving the fidelity of scene flow estimations in applications like 3D scene reconstruction and motion analysis.
\subsection{Occlusion-aware Cost Volume}
\begin{figure}[h]
\centering
\includegraphics[width=0.95\linewidth]{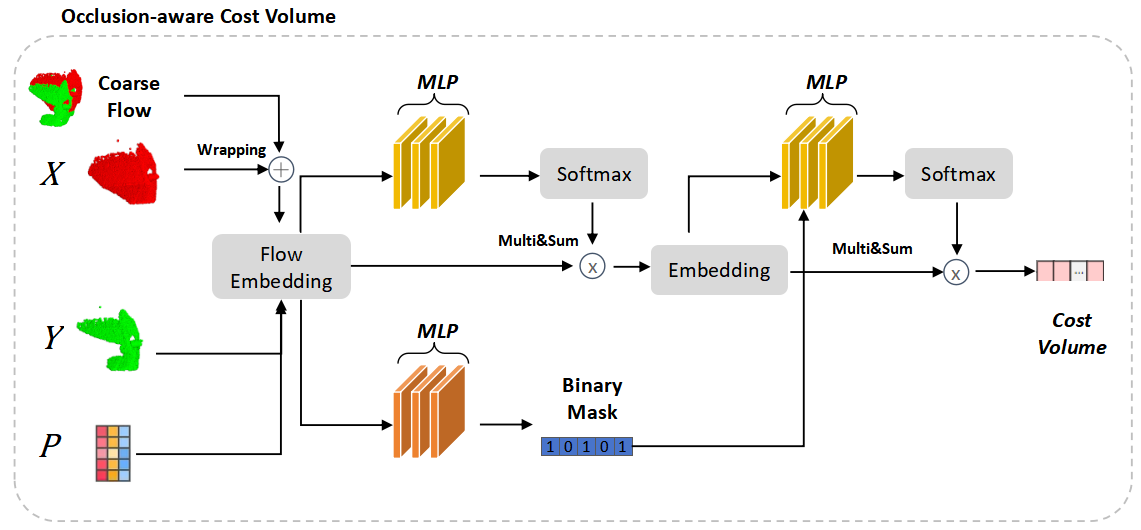}
\caption{Overview of Occlusion-aware Cost Volume (OCV) module.}

\label{fig:OCV}
\end{figure}
Cost Volume, in Figure.~\ref{fig:OCV}, is a measure of correlation between two frames of the point cloud. Previous work has confirmed its validity. We introduce an attention mechanism that includes occlusion information based on past models. Taking the computation of cost volume in layer $l$ as an example, we proceed to describe its computation steps.

The inputs are the coordinates and features of point cloud $P$ and $Q$ in layer $l$. Otherwise, we added the occlusion $O(x_i)$ as input to capture information in different dimensions. We first calculate the occlusion formula.
\begin{equation}
\begin{aligned}
    &cost(x^l_i,y^l_j) = MLP(x^{l}_i-y^{l}_j||p^{l}_i||q^{l}_j) \\
    &O^l(x_i) = Sigmoid(\underset{||x^{l}_i-y^{l}_j||<r}{MAX}\{cost(x^l_i,y^l_j)\})
    \label{eq7}
\end{aligned}
\end{equation}

Next, we learn the point cloud matching information for two frames. Inspired by \cite{wang2021hierarchical}, we compute the two-frame point cloud cost volume and then perform weighted average pooling for this data in $P^l$, which is to reduce the error due to long distance matching.

\begin{equation}
\begin{aligned}
    CV_1(x^l_i) = \underset{||x^{l}_i-y^{l}_j||<r}{\sum} Weight_1(x^l_i,y^l_j)*MLP(cost(x^l_i,y^l_j)) \\
    CV(x^l_i)  = \underset{||x^{l}_i-x^{l}_j||<r}{\sum}Weight_2(x^l_i,x^l_j,O^l_i)*MLP(CV_1(x^l_j))
    \label{eq8}
\end{aligned}
\end{equation}

Where \(Weight_1\) and \(Weight_2\) denote the weights function of each of the $N$ proximity points about point $xi$. As before, $||$ denotes matrix contact operation. For simplicity, $O^l_i$ represents $O(x_i)$.
\subsection{Flow Refinement}
We compute the obtained cost volume value, which is used to refine the previously obtained upsampled flow \(f^{l}_{u,i}\). In addition to the cost volume, we add the upsampled flow \(f^{l}_{u,i}\), the occlusions \( O(x_i)\), and the feature information \(p^l_i\). Unlike \cite{wang2021hierarchical,wang2022matters}, we discard using flow encoding as the information and add the occlusion information, and experiments show that our module reduces the computational cost while improving the accuracy.

\begin{equation}
\begin{aligned}
    \Delta f^l_i =& MLP(CV(x^l_i)||p^l_i||f^{l}_{u,i}||O(x_i)) \\
    &f^l_i = f^{l}_{u,i} + \Delta f^l_i
    \label{eq9}
\end{aligned}
\end{equation}

\subsection{Loss Fuction}
At each layer, we can obtain the estimated occlusion $O(x^l_i)$ and flow $f^l_i$. We adopt a cyclic strategy to compute the loss function for each layer and attach appropriate weights. We divide the loss into two components, the occlusion loss as well as the flow loss which is similar to \cite{ouyang2021occlusion}.

\begin{equation}
\begin{aligned}
    &Loss_o = \sum_{l=0}^{3}\beta ^l * \|O(x^l_i)-O_{gt}(x^l_i)\|_2 \\
    &Loss_f = \sum_{l=0}^{3}\beta ^l * \|f^l_i-f^l_{gt,i}\|_2 \\
    &Loss = \alpha * Loss_f + (1-\alpha)*Loss_o
    \label{eq10}
\end{aligned}
\end{equation}

Where $Loss_o$ and $Loss_f$ represent the occlusion loss and flow loss respectively.

\section{Experiment}
\begin{figure*}[!ht]

  \centering
  \centerline{\includegraphics[width=\textwidth]{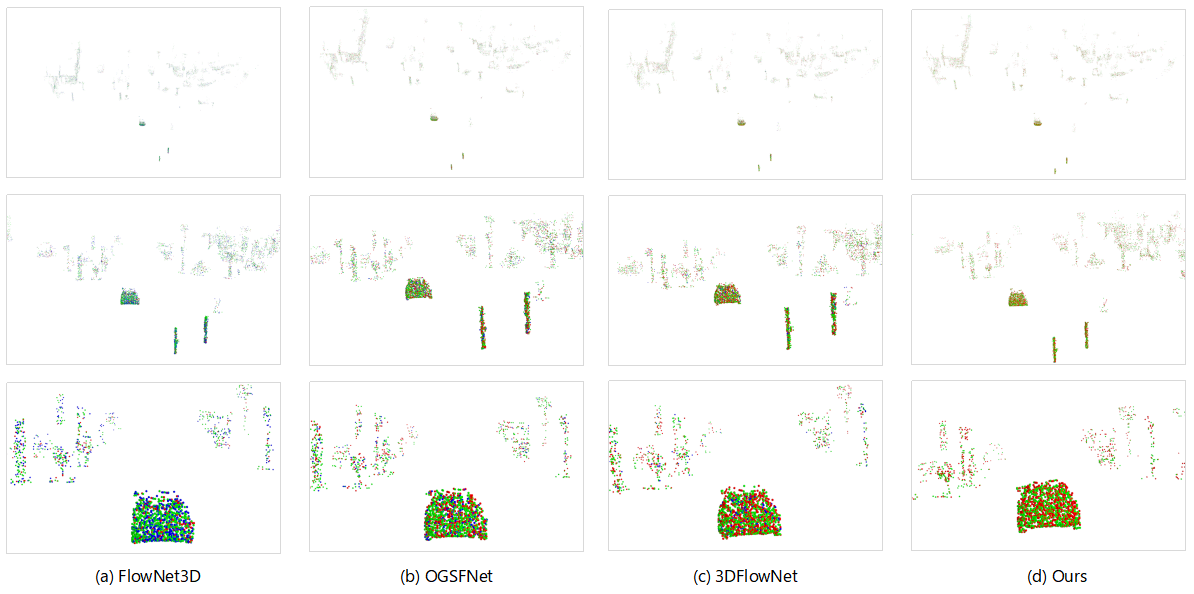}}
%

\caption{Visualization on KITTI\textsubscript{o}, red points represent the position at time $t$ wrapped by predicted flow, green represent the position at time $t+1$, and blue points indicate inaccurate predictions (measured by Acc3D Strict).}
\label{fig:vis}
\end{figure*}

\subsection{Experimental Setups}

\textbf{Dataset} We conduct our experiments on FT3D\textsubscript{o}\cite{mayer2016large} and KITTI\textsubscript{o}\cite{menze2015object,menze2018object} respectively. FlyingThings3D \cite{mayer2016large} is a synthetic dataset for optical flow, disparity and scene flow estimation. It consists of everyday objects flying along randomized 3D trajectories. In the field of point cloud scene flow estimation, there are two commonly used data processing methods. The first version is prepared by HPLFlowNet \cite{gu2019hplflownet}, we denote these datasets without occluded points as FT3D\textsubscript{s}. The second version is prepared by Flownet3D \cite{liu2019flownet3d}, we denote this occluded dataset as FT3D\textsubscript{o}. KITTI is a real-world scene flow dataset with 200 pairs for which 142 are used for testing without any fine-tuning. The KITTI can also be divided into occluded and non-occluded versions, named KITTI\textsubscript{s} and KITTI\textsubscript{o}, respectively. To verify the effectiveness of our model in occluded scenes, we take the processing in Flownet3D \cite{liu2019flownet3d} to generate the occluded datasets. 

\textbf{Details} Our model is trained based on pytorch, using NVIDIA GeForce RTX 3090 as the hardware device. we train our model on synthetic FT3D\textsubscript{o} training data and evaluate it on both FT3D\textsubscript{o} test set and KITTI\textsubscript{o} without finetune. Referring to most practices in the domain, we randomly take 8192 points per batch in training. In terms of model parameters, we set the upsampling range $N=32$ and the hyperparameters $\beta^l$ as [0.02, 0.04, 0.08, 0.16], $\alpha$ as 0.8 for training.  We set the learning rate to 0.001 and the decay factor to 0.5, and decay in every 80 training epochs.  We take Adam as the optimizer with default values for all parameters. In total, we train about 400 epochs.

\textbf{Evaluation Metrics} We test our model with four evaluation metrics, including End Point Error (EPE), Accuracy Strict (AS), Accuracy Relax (AR), and Outliers (Out). We denote the estimated scene flow and ground truth scene flow as $F$ and $F_{gt}$, respectively. EPE(m): $\|F-F_{gt}\|_2$ averaged over all points. AS: the percentage of points whose EPE $<$0.05m or relative error$<$5\%. AR: the percentage of points whose EPE $<$0.1m or relative error $<$10\%. Out: the percentage of points whose EPE $>$ 0.3m or relative error $>$ 10\%.
\begin{table}[!ht]
    \centering
    \caption{Comparison of our model with previous methods on the occluded datasets FT3D\textsubscript{o}. In these models, DELFlow uses a multimodal training approach where they use point clouds and images as input.}
    \begin{tabular}{l|l|l|l l l l}
    \hline
        Method & sup. & EPE↓ & AS↑ & AR↑ & Out↓  \\ \hline
        FlowNet3D & full & 0.169 & 0.254 & 0.579 & 0.789  \\ 
        FLOT & full & 0.156 & 0.343 & 0.643 & 0.700  \\   
        PointPWC-Net & full & 0.155 & 0.416 & 0.699 & 0.639  \\
        OGSFNet  & full & 0.122 & 0.552 & 0.777 & 0.518  \\ 
        FESTA & full & 0.111 & 0.431 & 0.744 & -  \\   
        FlowFormer & full & 0.077 & 0.720 & 0.866 & 0.316 \\
        CamLiRAFT & full& 0.076 &0.794 &0.904 &0.279  \\
        3DFlowNet & full & 0.063 & 0.791 & 0.909 & 0.279  \\ 
        DELFlow  & multi &0.058 &0.867 &0.932& - \\ \hline
        PointPWC-Net & self & 0.382 & 0.049 & 0.194 & 0.974  \\  \hline
        \textbf{Ours} & full & \textbf{0.054} & \textbf{0.828} & \textbf{0.922} & \textbf{0.223} \\ 
        \hline
    \end{tabular}
    \label{tb1}
\end{table}

\begin{table}[!ht]
    \centering
    \caption{Comparison of our proposed method with previous methods on the occluded datasets KITTI\textsubscript{o}. As previous method do, we train on FT3D\textsubscript{o} and test on KITTI\textsubscript{o} without any finetune. }
    \begin{tabular}{l|l|l|l l l l}
    \hline
        Method & sup. & EPE↓ & AS↑ & AR↑ & Out↓  \\ \hline 
        FlowNet3D & full & 0.173& 0.276& 0.609& 0.649  \\ 
        FLOT & full & 0.110 & 0.419 & 0.721 & 0.486 \\ 
        PointPWC-Net & full & 0.118& 0.403& 0.757 &0.497 \\
        OGSFNet & full &0.075& 0.706&0.869&0.328 \\
        FESTA & full &0.094 &0.449 &0.834 &-  \\ 
        FlowFormer & full & 0.074 & 0.784 & 0.883 & 0.262 \\
        3DFlowNet & full & 0.073 &0.819& 0.890 &0.261  \\ \hline
        PointPWC-Net & self & 0.337& 0.053& 0.213 &0.935  \\ 
        Self-Point-Flow & self & 0.105 &0.417& 0.725& 0.501   \\ \hline
        \textbf{Ours} & full & \textbf{0.065} & \textbf{0.856} & \textbf{0.911} & \textbf{0.221}  \\
        \hline
    \end{tabular}
    \label{tb1_2}
\end{table}
\subsection{Performance}
We train our model in FT3D\textsubscript{o}, at the same time we tested on the FT3D\textsubscript{o} test set and KITTI\textsubscript{o}, the results are shown in Table \ref{tb1}. We compare our performance with mainstream models in recent years, and our approach outperforms all other methods (Note that we compare under the occluded dataset). Our network improve performance by about 55.7\% compared to OGSFNet\cite{ouyang2021occlusion}, a previous model for occluded environments. Otherwise, we surpass 3DFlownet\cite{wang2022matters} by about 14.3\%, which is the optimal algorithm in recent years. Moreover, in a comparison with the multimodal method DELFlow , we were able to exceed its performance when using only the point cloud as input. Also, the experimental results on KITTI prove that our model has generalisation ability. On EPE3D, we outperform 3DFlownet\cite{wang2022matters} 0.008. And there is a considerable improvement in AS and AR, compared to the previous advanced method, we improve performance by about 4.5\% and 2.4\%. On Outliers , we reduce the error rate by about 3\% compared with the previous one, and such results show that our model achieves a leading level in error control.

For the occlusion estimation, our model achieves a 93.6\% accuracy on the FT3D\textsubscript{o}, it shows that our model can effectively perceive the occlusion and avoid the error generated by the occlusion, which is one of the reasons for the efficient performance of our model.

\subsection{Ablation Study}

To validate the effectiveness of our key components, we perform ablation experiments on two proposed modules. First we replace the CMU(Correlation Matrix Upsampling) based upsampling algorithm with the most common trilinear upsampling. Secondly, we use the Cost Volume module, which also integrates occlusion prediction, to compare with our OCV(Occlusion-aware Cost Volume). We replace OCV in the model with Cost Volume in \cite{ouyang2021occlusion} and test it. We sequentially demonstrate the effectiveness of CMU and OCV by arranging and combining the modules of the existing model with the traditional approach of the past. We train the replaced model and the results are shown in table \ref{tb3}.

\begin{table}[!ht]
    \centering
    \caption{The ablation experiment focuses on the two proposed modules. We replace the CMU module with  trilinear interpolation upsampling and the OCV using the Cost Volume calculation method in \cite{ouyang2021occlusion}. The experimental results show the effectiveness of our modules.}
    \begin{tabular}{c|c c|l l l l}
    \hline
        Data. & CMU & OCV &  EPE↓ & AS↑ & AR↑ & Out↓  \\ \hline 
        \multirow{4}{1cm}{\centering Fly.} &  &  & 0.065 & 0.777 & 0.905 & 0.301  \\ 
          & & \ding{51} &  0.061 & 0.790 & 0.914 & 0.290   \\
         &\ding{51} &  &   0.057 & 0.816 & 0.920 & 0.240  \\  
         &\ding{51} & \ding{51} & 0.054 & 0.828 & 0.922 & 0.223 \\ \hline
        \multirow{4}{1cm}{\centering KI.} &  &  & 0.807 & 0.810 & 0.881 & 0.263  \\ 
        & & \ding{51} &   0.077 & 0.818 & 0.883 & 0.252  \\
         & \ding{51} & & 0.072 & 0.849 & 0.904 & 0.226  \\ 
        &\ding{51} & \ding{51} & 0.065 & 0.856  & 0.911 & 0.221 \\
        \hline
        \end{tabular}
    \label{tb3}
\end{table}

As shown in Table \ref{tb3}, when we discarded two modules, the metrics declined to varying degrees. First, when we do not use CMU and OCV, EPE3D rises to 0.065, and the accuracy of each is reduced by 2-4\%. When we introduced OCV, EPE3D was elevated by about 6\%, and accuracy AS and AR also improved somewhat. When we added CMU, the model performance was improved compared to the past, with EPE3D, AS,AR, improved by 12.3\%, 5\%, and 1.5\%, respectively. The best performance can be obtained when we add two modules at the same time. The same result can be seen on KITTI dataset. Although the degree of decrease in epe3d is not significant, we have a substantial improvement in both AS and AR due to the effect of the finer upsampling, which shows that our module is still valid in the real-world dataset, as well.

\begin{figure}[h]
\centering
\includegraphics[width=0.7\linewidth]{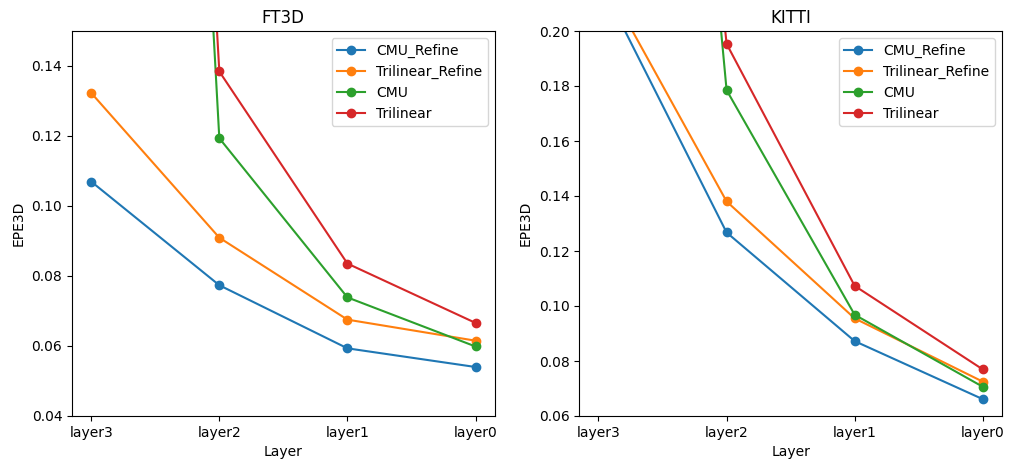}
\caption{The EPE values of the predicted flow vectors for each layer are shown in Fig. The two lines represent the epe values of the flows obtained by CMU(Correlation Matrix Upsampling) and  trilinear interpolation upsampling, respectively. From the figure, it can be seen that the up-sampled flow obtained through CMU has higher accuracy.}

\label{fig:chart1}
\end{figure}
To further illustrate the role of the CMU in upsampling, we measure the error of the flow computed by each layer. Again we use a traditional linear method and CMU to compare the accuracy of both methods before and after flow refinement. The graphs illustrate that our module improves the accuracy of the streams at all layers and can be integrated into any model from coarse to fine.

\subsection{Sampling Range Setting}

\begin{table}[!ht]
    \centering
    \caption{The graphs illustrate the accuracy of the model at different sampling ranges in FT3D dataset.}
    \begin{tabular}{l| l l l l l}
    \hline
          N &  EPE↓ & AS↑ & AR↑ & Out↓  \\ \hline 
          8 & 0.059& 0.813 & 0.918 & 0.249   \\
          16 & 0.056& 0.819 & 0.920 & 0.236  \\  
          32 & 0.054& 0.828 & 0.922 & 0.223 \\ 
          64 & 0.059& 0.794 & 0.912 & 0.266 \\
           \hline
        
        \end{tabular}
    \label{tb4}
\end{table}
In this subsection we explore the effect of the value of the number of samples N in cmu on the final results. If we use a smaller range, it may lead to errors due to occlusion, and if the range is larger it will lead to a decrease in computational power and accuracy. We train on flying to explore the change in EPE3D for each layer and the final results when only the sampling range is changed.

\begin{figure}[h]
\centering
\includegraphics[width=0.8\linewidth]{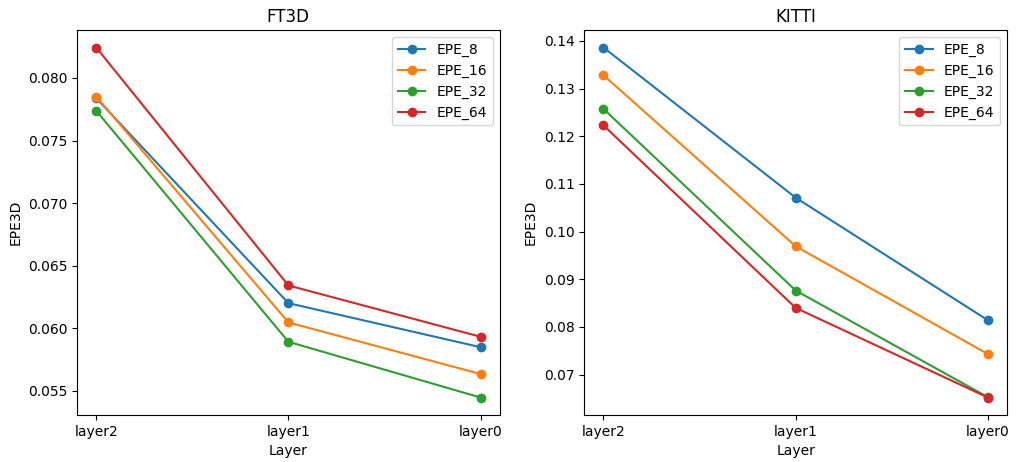}
\caption{The line graph illustrates the EPE3D values for each layer of the network structure when $N=8,16,32,64.$}

\label{fig:chart2}
\end{figure}

From the graphs and tables, the effect on the accuracy of the model at different N settings can be obtained. We conducted experiments in FT3D and KITTI respectively. On the FT3D dataset, the best accuracy is possessed when set to $N = 32$, and a large drop in accuracy occurs when $N = 8$ and 64 due to too small and too large sensory fields. On the KITTI dataset, $N=64$ has the best accuracy, this is because in real scene datasets the points have a large distance to move and when we expand the receptive field it gives better results. Combining the performance of the above two datasets, we take $N=32$, partly because we can get better model accuracy and partly because the time loss is smaller.

\subsection{Assessment of Model Efficiency}
The efficiency of a model has been seen as an important part of evaluating model performance in recent years. Due to the limitation of equipment performance, how to make the model lightweight has also received much attention. We compare the model size and run time with the current mainstream models to show that the efficiency of our model is within a reasonable range. All models were experimented on the same hardware.
\begin{table}[!ht]
    \centering
    \caption{Evaluation of model size and run time.}
    \small
    \begin{tabular}{l| c c c c c}
    \hline
          Metrics &  FlowNet3D & HPLFlowNet & FESTA & 3DFlownet & \textbf{ours}  \\ \hline 
          Size (MB) & 14.9& 231.8 & 16.1 & 19 & 22  \\
          Time (ms) & 34.9& 93.1 & 67.8 & 60.1 & 64.5 \\
         \hline
        \end{tabular}
    \label{tb5}
\end{table}

We select \cite{liu2019flownet3d,gu2019hplflownet,wang2021festa,wang2022matters} as a comparison to our model. From table \ref{tb5} we can see that the efficiency of our model is better than HPLFlowNet \cite{gu2019hplflownet} and slightly lower than the remaining ones. However, our accuracy is large higher than these methods, and the appropriate sacrifice of efficiency is acceptable.

\section{Conclusion}
In this study, we address the challenge of robustly matching successive frames in point cloud sequences. Despite the recognized potential of neural network-based scene flow estimation, its application in occlusion-rich environments remains partially explored. To advance this area, we introduce the Correlation Matrix based Upsampling Flownet (CMU-Flownet), that seamlessly integrates an occlusion estimation module within its cost volume layer via an Occlusion-aware Cost Volume (OCV) mechanism. Additionally, our model incorporates a novel upsampling strategy utilizing a Correlation Matrix to evaluate point-level similarity. Through rigorous empirical evaluations on datasets known for their occluded scenarios, such as Flyingthings3D and Kitti, CMU-Flownet demonstrates superior performance over existing methods across various metrics.

\section{Acknowledge}
The paper is supported by the Natural Science Foundation of China (No. 62072388), Collaborative Project fund of Fuzhou-Xiamen-Quanzhou Innovation Zone(No.3502ZCQXT202001),the industry guidance project foundation of science technology bureau of Fujian province in 2020(No.2020H0047), , and Fujian Sunshine Charity Foundation.
\bibliography{sample-base}

\end{document}